\documentclass[10pt,twocolumn,letterpaper]{article}

\usepackage{iccv}
\usepackage{times}
\usepackage{epsfig}
\usepackage{graphicx}
\usepackage{amsmath}
\usepackage{amssymb}

\usepackage{multirow}
\usepackage{multicol}
\usepackage{lineno}
\usepackage{algorithm}
\usepackage{algorithmic}
\usepackage{booktabs}
\usepackage{caption}

\usepackage{bbding}
\usepackage[breaklinks=true,bookmarks=false,colorlinks]{hyperref}

\iccvfinalcopy 

\def\iccvPaperID{****} 

\ificcvfinal\fi

\begin{document}
\title{MasaCtrl: Tuning-Free \underline{M}utu\underline{a}l \underline{S}elf-\underline{A}ttention \underline{Control} for Consistent \\Image Synthesis and Editing}

\author{
\vspace{0.2cm}
Mingdeng~Cao$^{1,2*}$\quad
Xintao~Wang$^2$\Envelope\quad
Zhongang~Qi$^2$\quad
Ying~Shan$^2$\quad
Xiaohu~Qie$^2$\quad
Yinqiang~Zheng$^1$\Envelope \\
$^1$The University of Tokyo \qquad $^2$ARC Lab, Tencent PCG \vspace{0.2cm} \\
\large\url{https://github.com/TencentARC/MasaCtrl}\vspace{-0.4cm}
}

\newcommand{\xt}[1]{{\color{red}{Xintao: #1}}}


\newcommand*{\method}{\mbox{MasaCtrl}}

\twocolumn[{%
  \renewcommand\twocolumn[1][]{#1}%
  \maketitle
  \begin{center}
   \centering
   \includegraphics[width=\textwidth]{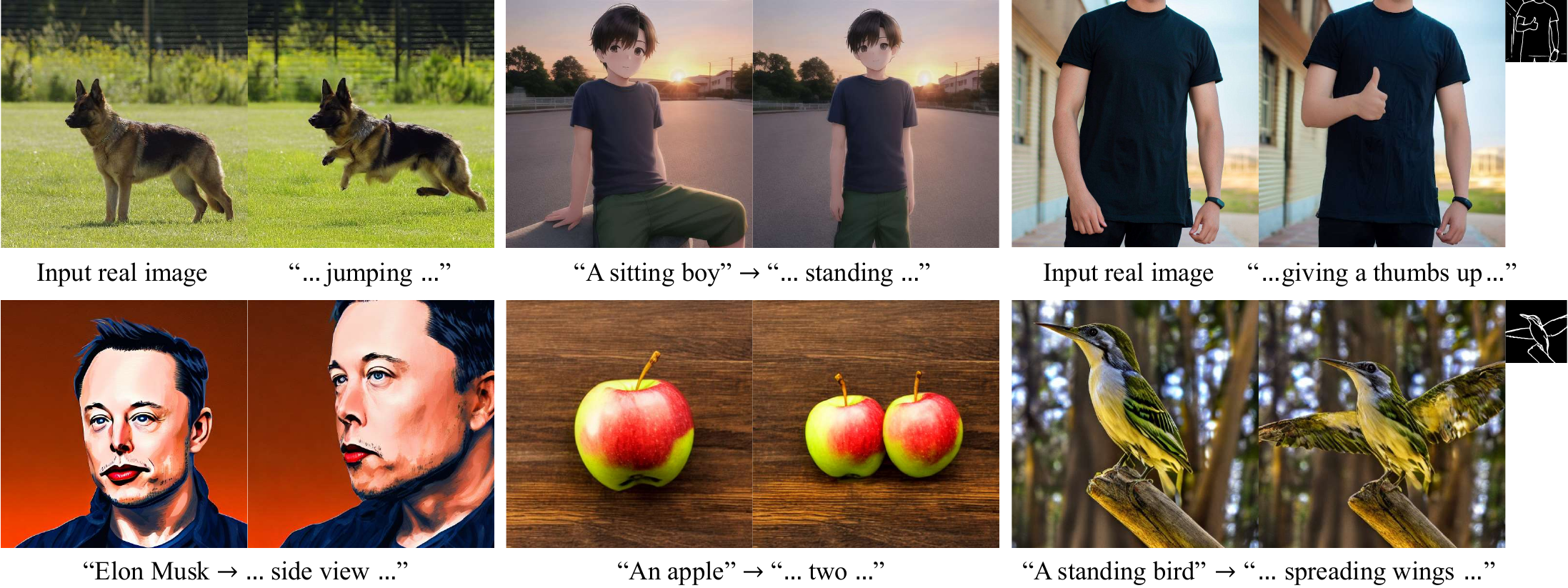}
   \captionof{figure}{Our method \textbf{MasaCtrl} can perform text-based non-rigid image synthesis and real image editing without finetuning. Meanwhile, our method can be easily integrated into controllable diffusion models, like T2I-Adapter~\cite{mou2023t2i} or ControlNet~\cite{zhang2023adding}, to perform more consistent and faithful synthesis and editing~(last column).
   }
  \end{center}%
 }]

\begin{abstract}
\let\thefootnote\relax\footnotetext{*Work done during an internship at ARC Lab, Tencent PCG.}
Despite the success in large-scale text-to-image generation and text-conditioned image editing, existing methods still struggle to produce consistent generation and editing results. For example, generation approaches usually fail to synthesize multiple images of the same objects/characters but with different views or poses. Meanwhile, existing editing methods either fail to achieve effective complex non-rigid editing while maintaining the overall textures and identity, or require time-consuming fine-tuning to capture the image-specific appearance. In this paper, we develop \textbf{MasaCtrl}, a tuning-free method to achieve consistent image generation and complex non-rigid image editing simultaneously. Specifically, MasaCtrl converts existing self-attention in diffusion models into mutual self-attention, so that it can query correlated local contents and textures from source images for consistency. To further alleviate the query confusion between foreground and background, we propose a mask-guided mutual self-attention strategy, where the mask can be easily extracted from the cross-attention maps. Extensive experiments show that the proposed MasaCtrl can produce impressive results in both consistent image generation and complex non-rigid real image editing.
\end{abstract}

\section{Introduction}
Recent advances in text-to-image (T2I) generation~\cite{ramesh2021zero, nichol2021glide, yu2022parti, ramesh2022hierarchical, rombach2022high} have achieved great success. Those large-scale T2I models, such as Stable Diffusion~\cite{rombach2022high}, can generate diverse and high-quality images conforming to given text prompts. When leveraging the T2I models, we can also perform promising text-conditioned image editing~\cite{nichol2021glide, hertz2022prompt, tumanyan2022plug, parmar2023zero}. However, there is still a large gap between our needs and existing methods in terms of \textit{consistent} generation and editing.

\begin{figure}[!t]
    \centering
    \includegraphics[width=\linewidth]{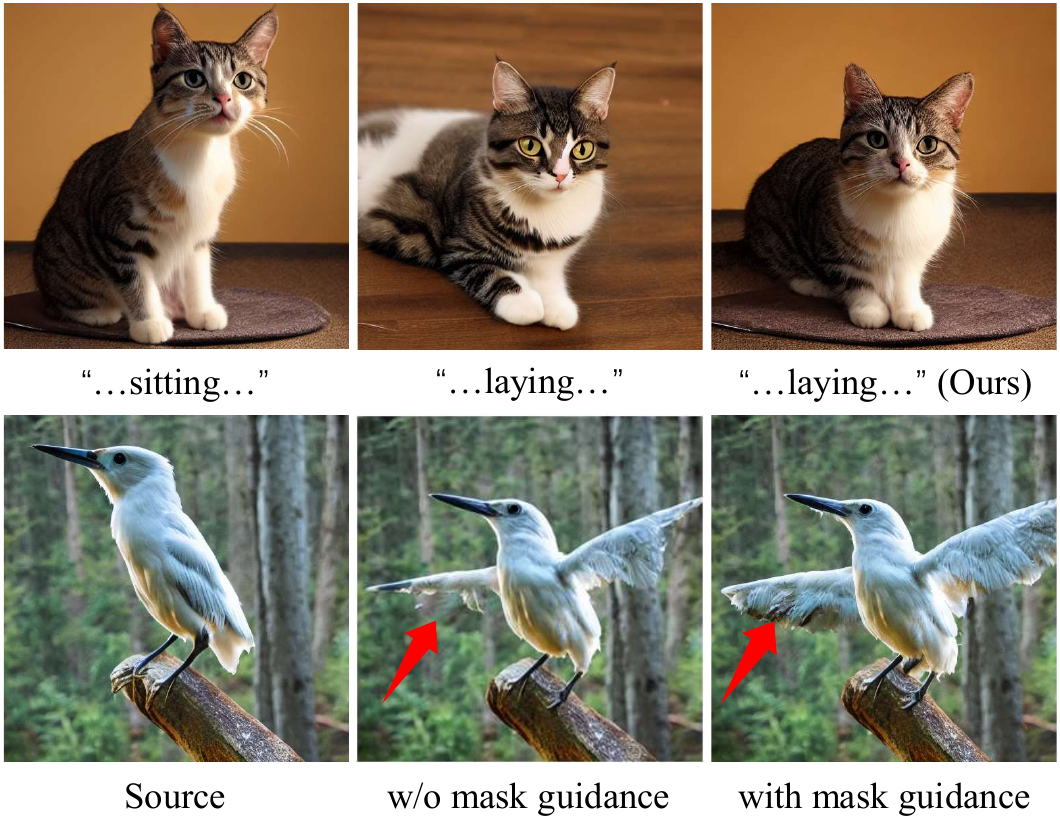}
    \caption{First row: images synthesized from fixed random seed~(middle, changed identity) and our method (right, maintained identity). Second row: image synthesized without mask guidance (middle) and with mask guidance (right).}
    \label{fig:motivation}
\end{figure}

For the text-to-image generation, we usually want to generate several images of the same objects/characters but with different views or complex non-rigid variances (\eg, the changes of posture). Such capabilities are urgently needed for creating comic books and generating short videos using existing powerful T2I image models. However, this requirement is highly challenging. Even if we fix the random input noise and use very similar prompts (\eg, `a sitting cat' \vs `a laying cat', see Fig.~\ref{fig:motivation}),  the two generated images are very different in both structures and identity.

 For text-conditioned image editing, existing methods~\cite{hertz2022prompt, tumanyan2022plug, parmar2023zero} achieve impressive editing effects in image translation, style transfer, and appearance replacement while keeping the input structure and scene layout unchanged. However, those methods usually fail to change poses or views while maintaining the overall textures and identity, leading to inconsistent editing results. The latter editing way is a more complicated \textit{non-rigid editing} for practical use.
Imagic~\cite{kawar2022imagic} is then proposed to address this challenge. It allows complex non-rigid edits while preserving its original characteristics. 
It can make a standing dog sit down, cause a bird to spread its wings, \etc.
Nevertheless, it requires fine-tuning the entire T2I diffusion model and then optimizing the textual embedding to capture the image-specific appearance for each edit, which is time-consuming and impractical for real-world applications.

In this paper, we aim to develop a \textit{tuning-free} method to address the above challenges, enabling a more consistent generation of multiple images and complex non-rigid editing without fine-tuning~\footnote{In fact, we can also regard the process of image editing as a kind of generating multiple consistent images simultaneously. We can invert one image and then perform consistent synthesis based on it with the pose, view, and non-rigid variance while maintaining the overall characteristics, textures, and identity. For simplicity, we mainly describe the synthesis process, but our method can be directly applied to editing.}.
The core question is how to keep consistent. Unlike previous works~\cite{hertz2022prompt, parmar2023zero, chefer2023attend} that usually operate on cross-attention in T2I models, we propose to convert existing \textit{self-attention} to mutual self-attention, so that it can query correlated local structures and textures from a source image for consistency.
Specifically, we first generate an image or invert a real image, resulting in the diffusion process (DP1) for the source image synthesis. 
In the new diffusion process (DP2) of generating a new image or editing an existing one, we can use the Query features in DP2 self-attentions to query the corresponding Key and Value features in DP1 self-attentions. 
In other words, we transform the existing self-attention into `cross-attention', where the crossing operation happens in the self-attentions of two related diffusion processes, rather than between the U-Net feature and text embeddings. We call this `crossing self-attention as \textit{mutual self-attention}. 
However, directly applying this strategy can only generate images almost identical to the source image and cannot comply with the target text prompt (as analyzed in Fig.~\ref{fig:step_layer_analysis}). Thus we further control the denoising time step and the layer position in U-Net for performing mutual self-attention to achieve consistent synthesis and editing. More analyses are in Sec.~\ref{sec:mutual-self-attention} and Sec.~\ref{sec:ablation_study}.
In this way, we can use contents in the source image as the generation material to better maintain the texture and identity. Meanwhile, its structure, pose, and non-rigid variances can be controlled by target text, or guided by recent controllable T2I-Adapters~\cite{mou2023t2i} or ControlNet~\cite{zhang2023adding}.

The proposed mutual self-attention control can work well for images with disentangled foreground and background, but often fails in the synthesis and editing process where the foreground and background have similar patterns and colors. In these cases, mutual self-attention tends to confuse the foreground and background, leading to a messy result (Fig.~\ref{fig:motivation}, 2nd row).
To address this problem, we further propose a mask-guided mutual self-attention. 
Specifically, we first utilize the cross attention in T2I diffusion models to extract a mask associated with the main object in the image. This mask can successfully separate foreground and background, and restrict the new foreground/background features to query only  foreground/background features of the source image, respectively.
Such an operation can effectively alleviate the query confusion between the foreground and background.

Our contributions can be summarized as follows.
\textbf{1)} We propose a tuning-free method, namely, MasaCtrl, to simultaneously achieve consistent image synthesis and complex non-rigid image editing. 
\textbf{2)} An effective mutual self-attention mechanism with delicate designs is proposed to change pose, view, structures, and non-rigid variances while maintaining the characteristics, texture, and identity.
\textbf{3)} To alleviate the query confusion between foreground and background, we further propose a masked-guided mutual self-attention, where the mask can be easily computed from the cross-attentions.
\textbf{4)} Experimental results have shown the effectiveness of our proposed MasaCtrl in both consistent image generation and complex non-rigid real image editing.

\section{Related Work}

\subsection{Text-to-Image Generation}
Early image generation methods conditioned on text description mainly based on GANs~\cite{reed2016generative, zhang2017stackgan, zhang2018stackgan++, xu2018attngan, li2019controllable, brock2018large, zhu2019dm, ye2021improving, zhang2021cross, tao2022df}, due to their powerful capability of high-fidelity image synthesis. These models try to align the text descriptions and synthesized image contents via multi-modal vision-language learning and have achieved cheerful synthesis results on domain-specific datasets. 
Text-to-image generation with auto-regressive and diffusion models has obtained impressive diversity results. DALL·E~\cite{ramesh2021zero}, CogView~\cite{ding2021cogview} and Parti~\cite{yu2022parti}, large-scale text-to-image models trained with a large amount of data, enable generating images from open-domain text descriptions. However, the auto-regressive generation nature attributes to the slow generation process defect. Most recently, diffusion models~\cite{song2019generative, ho2020denoising, nichol2021improved, dhariwal2021diffusion} have shown superior generative power and achieved state-of-the-art synthesis results in terms of image quality and diversity than previous GAN-based and auto-regressive image generation models. By conditioning the text prompt into the diffusion model, various text-to-image diffusion models GLIDE~\cite{nichol2021glide}, VQ-Diffusion~\cite{gu2022vector}, LDM~\cite{rombach2022high}, DALL·E 2~\cite{ramesh2022hierarchical}, and Imagen~\cite{saharia2022photorealistic} have been developed. They can synthesize high-quality images that highly comply with the given text description.

\subsection{Text-guided Image Editing. }
Text-guided image editing is a challenging task that aims to manipulate images according to natural language descriptions. Previous methods based on generative adversarial networks (GANs)~\cite{nam2018text, li2020manigan, xia2021tedigan, patashnik2021styleclip} have achieved some success on domain-specific datasets (\eg, face datasets), but they have limited applicability and generality. A recent approach based on auto-regressive models, VQGAN-CLIP~\cite{crowson2022vqgan}, combines VQGAN~\cite{esser2021taming} and CLIP~\cite{radford2021learning} to produce high-quality images and precise edits with diverse and controllable results. However, this approach suffers from slow generation speed and high computational cost.

Different from previous methods based on GANs or auto-regressive models, diffusion models offer a fast and efficient way to synthesize and edit images conditioned on text prompts. However, existing diffusion-based methods have some limitations in terms of local and global editing. For example, the works~\cite{nichol2021glide, avrahami2022blended} require extra masks to edit local regions of the image; ~\cite{kim2022diffusionclip} can edit global aspects of the image by changing the text prompt directly, but cannot modify local details; ~\cite{hertz2022prompt,tumanyan2022plug} use cross-attention or spatial features to edit both global and local aspects of the image by changing the text prompt directly, but they tend to preserve the original layout of the source image and fail to handle non-rigid transformations (\eg, changing object pose). In contrast, we propose a novel approach that leverages the self-attention mechanism to achieve consistent and complex non-rigid image synthesis and editing. Our approach can modify various object attributes (\eg, pose, shape, color) by changing the text prompt accordingly. The most related work to ours is Imagic~\cite{kawar2022imagic}, which also enables various non-rigid image editing by changing the prompts directly. However, unlike our approach which can edit images on the fly, Imagic requires careful optimization of the textual embedding and fine-tuning of the model, which is time-consuming and unfriendly for ordinary users.

\begin{figure*}
    \centering
    \includegraphics[width=0.99\linewidth]{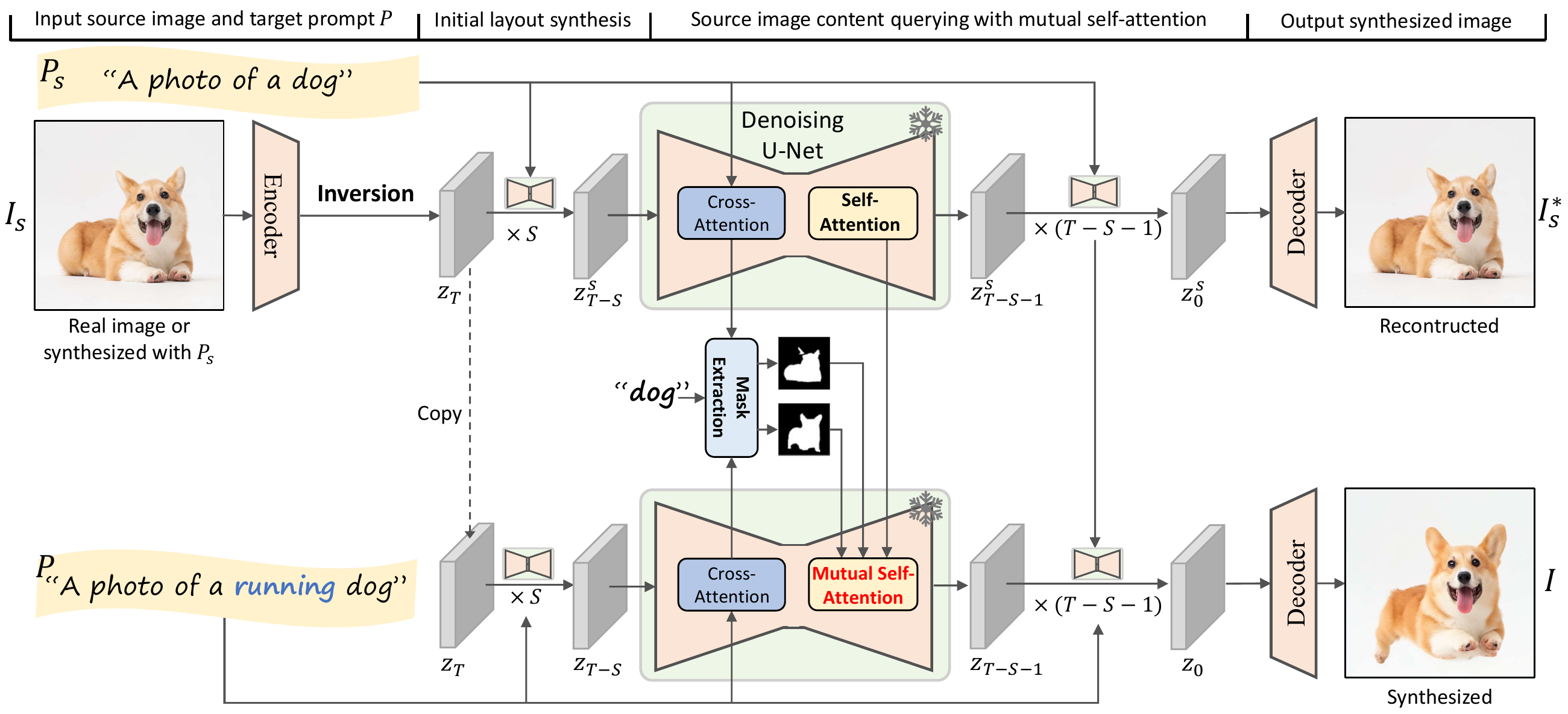}
    \caption{Pipeline of the proposed MasaCtrl. Our method tries to perform complex non-rigid image editing and synthesize content-consistent images. The source image is either real or synthesized with source text prompt $P_s$. During the denoising process for image synthesis, we convert the self-attention into mutual self-attention to query image contents from source image $I_s$, so that we can synthesize content-consist images under the modified target prompt $P$. } 
    \label{fig:main_arch}
\end{figure*}

\section{Preliminaries}
\subsection{Latent Diffusion Models}
Diffusion models~\cite{ho2020denoising, song2020denoising, nichol2021improved} are generative models that can synthesize desired data samples from Gaussian noise via iteratively removing the noise. 
A diffusion model defines a forward process and a corresponding reverse process. The forward process adds the noise to the data sample $x_0$ to generate the noisy sample $x_t$ with a predefined noise adding schedule $\alpha_t$ at time-step $t$:
\begin{equation}
    \label{eq:diffsuion}
    q(x_t|x_0) = \mathcal{N}(x_t; \sqrt{\bar{\alpha}}, (1 - \bar{\alpha})_t I),
\end{equation}
where $\bar\alpha_t = \prod_{i=1}^{t}\alpha_i$. After $T$ steps, the data sample $x_0$ is transformed into Gaussian noise ${x}_T \sim \mathcal{N}(0, 1)$. The reverse process tries to remove the noise and generate a clean sample $x_{t-1}$ from the previous sample $x_t$:
\begin{equation}
    \label{eq:denoise}
    p_\theta(x_{t-1}|x_t) = \mathcal{N}(x_{t-1}; \mu_{\theta}(x_t, t), \sigma_t),
\end{equation}
where $\mu_\theta$ and $\sigma_t$ are the corresponding mean and variance. The variance is a time-dependent constant, and the mean $\mu_\theta(x_t, t) = \frac{1}{\sqrt{\alpha_t}}(x_t - \epsilon\frac{1-\alpha_t}{1-\bar\alpha_t})$ can be solved by using a neural network $\epsilon_\theta(x_t, t)$ to predict the noise $\epsilon$. To train such a noise estimation network $\epsilon_\theta$, the object is a simplified mean-squared error:
\begin{equation}
    \label{eq:objective}
    \mathcal{L}_{\text{simple}} = \mathrm{E}_{x_0, \epsilon, t}(\| \epsilon - \epsilon_\theta(x_t, t) \|).
\end{equation}
Therefore, by sampling $x_{t-1}$ iteratively, a data sample $x_0$ can be synthesized from random Gaussian noise $x_T$. In addition, text prompts $P$ can be conditioned into the predicted noise $\epsilon_\theta(x_t, t, P)$ so that the diffusion models can synthesize text-complied images. 

Our method is based on the recent state-of-the-art text-conditioned model Stable Diffusion (SD)~\cite{rombach2022high}, which further performs the diffusion-denoising process in the latent space rather than image space. A pretrained image autoencoder network encodes the image into latent representations $z$.
We then train the denoising network $\epsilon_\theta$ in the latent space. After training, we can sample a random noise $z_T \sim \mathcal{N}(0, 1)$ and perform the latent denoising process. The denoised latent representation $z_0$ can be decoded into an image using the pretrained autoencoder. The structure of the denoising backbone $\epsilon_\theta$ is realized as a time-conditional U-Net~\cite{ronneberger2015unet}.

\subsection{Attention Mechanism in Stable Diffusion}
The denoising U-Net $\epsilon_\theta$ in the SD model, consists of a series of basic blocks, and each basic block contains a residual block~\cite{he2016deep}, a self-attention module, and a cross-attention~\cite{vaswani2017attention} module. At denoising step $t$, the features from the previous $(l{-}1)$-th basic block first pass through the residual block to generate intermediate features $f^l_t$; then they are reorganized by a self-attention layer, and receive textual information from the given text prompt $P$ by the following cross-attention layer. The attention mechanism can be formulated as follows:
\begin{equation}
    \label{eq:attention}
    \text{Attention}(Q, K, V) = \text{Softmax}(\frac{QK^T}{\sqrt{d}})V,
\end{equation}
where $Q$ is the query features projected from the spatial features, and $K, V$ are the key and value features projected from the spatial features (in self-attention layers) or the textual embedding (in cross-attention layers) with corresponding projection matrices. $A = \text{Softmax}(\frac{QK^T}{\sqrt{d}})$ is the attention map used to aggregate the value $V$. 

These attention layers in the SD model contain much information for the overall structure/layout and content formation of the synthesized image~\cite{hertz2022prompt, tumanyan2022plug}. The internal cross-attention maps are high-dimensional tensors that bind the spatial pixels and text embedding extracted from the prompt text~\cite{hertz2022prompt}, and they are explored for image editing~\cite{hertz2022prompt} and faithful image synthesis~\cite{chefer2023attend}. Meanwhile, the features in the self-attention layer are employed as plug-and-play features to be injected into specified layers in U-Net to perform image translation. However, these controls cannot perform non-rigid editing~(\eg, pose change) since they maintain the semantic layout and structures. 
Inspired by the phenomenon that performing self-attention across batches can generate similar image contents, which is also observed in Tune-A-Video~\cite{wu2022tune}, we adapt the self-attention mechanism in T2I model to query contents from source images with delicate designs. Thus we can perform various consistent synthesis and non-rigid editing that change the layout and structure of the source image while preserving image contents.

\section{Tuning-Free Mutual Self-Attention Control}
Given a source image $I_s$ and the corresponding text prompt $P_s$, our goal is to synthesize the desired image $I$ that complies with the target edited text prompt $P$ (directly modified from $P_s$). Note that the edited target image $I$ is spatially edited from $I_s$ and should preserve the object contents (\eg, textures and identity) in $I_s$. For instance, consider a photo (corresponding to $P_s$) where a dog is sitting, and we want the dog to be in the running pose with the edited text prompt $P$ (by adding the `running' into the source prompt $P_s$) (see Fig.~\ref{fig:main_arch}). This task is highly challenging, and previous diffusion-based tuning-free methods can hardly handle it~\cite{hertz2022prompt, tumanyan2022plug}. Directly utilizing $P$ for the synthesis will generate an image $I^*$ complied with the input prompt. Nevertheless, the objects in it are probably different from the ones in source image $I_s$~(see an example in Fig.~\ref{fig:motivation}), even with the fixed random seeds~\cite{hertz2022prompt}. 

Our core idea is to combine the semantic layout synthesized with the target prompt $P$ and the contents in the source image $I_s$ to synthesize the desired image $I$. 
To achieve so, we propose MasaCtrl, which adapts the self-attention mechanism in the SD model into a cross one to query semantically similar contents from the source image. Therefore, we can synthesize the target image $I$ by querying the contents from $I_s$ with the modified self-attention mechanism during the denoising process. We can achieve so for the following reasons: 1) the image layout is formed in the early denoising steps (shown in Fig.~\ref{fig:intermediate_vis}(a)); 2) in addition, as shown in Fig.~\ref{fig:intermediate_vis}(b), the encoded query features in the self-attention layer are semantically similar (\eg, the horses are in the same color), thus one can query content information from another.

The overall architecture of the proposed pipeline to perform synthesis and editing is shown in Fig.~\ref{fig:main_arch}, and the algorithm is summarized in Alg.~\ref{alg:masactrl}. The input source image $I_s$ is either a real image or a generated one from the SD model with text prompt $P_s$~\footnote{When $I_s$ is a real image, we set the text prompt $P_s$ as null and utilize the deterministic DDIM inversion~\cite{song2020denoising} to invert the image into a noise map.}. During each denoising step $t$ of synthesizing target image $I$, we assemble the inputs of the self-attention by \textbf{1)} keeping the current Query features $Q$ unchanged, and \textbf{2)} obtaining the Key and Value features $K_s$, $V_s$ from the self-attention layer in the process of synthesizing source image $I_s$. We dub this strategy mutual self-attention, and more details are in Sec.~\ref{sec:mutual-self-attention}.

Meanwhile, we also observe the edited image often suffers from the problem of confusion between the foreground objects and background. Thus, we propose a mask-aware mutual self-attention strategy guided by the masks obtained from the cross-attention mechanism. The object mask is automatically generated from the cross-attention maps of the text token associated with the foreground object. Please refer to Sec.~\ref{sec:mask-guided} for more details.

In addition, since the edited prompt $P$ may not yield desired spatial layouts due to the inner limitations of the SD model, our method MasaCtrl can be easily integrated into existing controllable image synthesis method~(\eg, T2I-Adapter~\cite{mou2023t2i} and ControlNet~\cite{zhang2023adding}) for more faithful non-rigid image editing. Please refer to Sec.~\ref{sec:controllable} for more details.

\subsection{Mutual Self-Attention} \label{sec:mutual-self-attention}
\begin{figure}[t]
    \centering
    \includegraphics[width=0.99\linewidth]{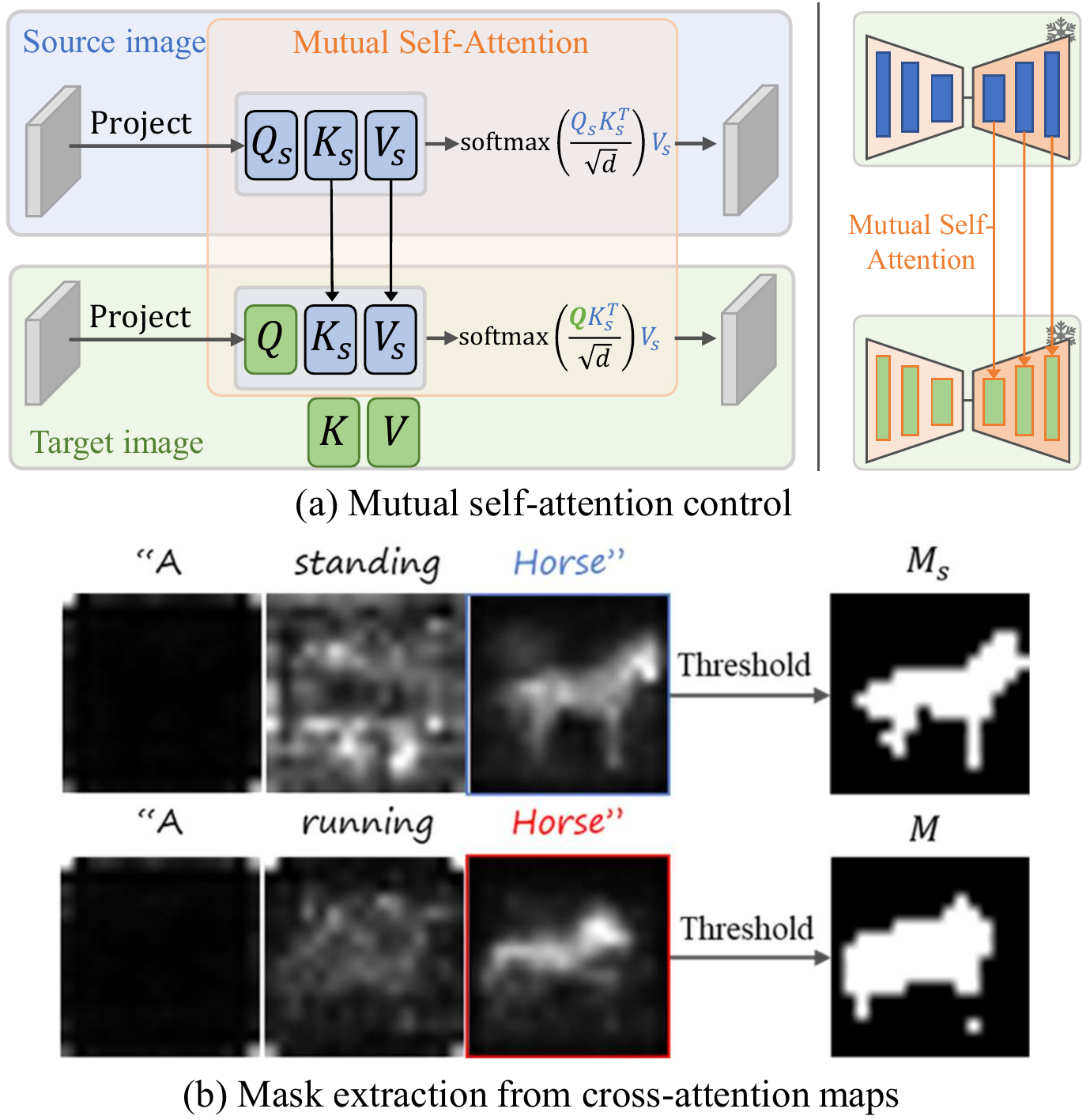}
    \caption{(a) The mutual self-attention control to query contents from the source image in the decoder part of denoising U-Net; and (b) mask extraction strategy from cross-attention maps.}
    \label{fig:mutual_selfattention}
\end{figure}

To obtain image contents from the source image $I_s$, we propose mutual self-attention, which converts the existing self-attention in T2I models into `cross-attention', where the crossing operation happens in the self-attentions of two related diffusion processes.
Specifically, as shown in the left part of Fig.~\ref{fig:mutual_selfattention}(a), at denoising step $t$ and layer $l$, the query features are defined as the projected query features $Q^l$ in the self-attention layer, and the content features are the key features $K^l_s$ and value features $V^l_s$ from the corresponding self-attention layer in the process of reconstructing the source image $I_s$. After that, we perform attention according to Eq.~\ref{eq:attention} to aggregate the contents from the source image.

\begin{algorithm}[!t]    
    \caption{MasaCtrl: Tuning-Free Mutual Self-Attention Control}
        \textbf{Input:} A source prompt $P_s$, a modified prompt $P$, the source and target initial latent noise maps $z^s_T$ and $z_T$. \\
        \textbf{Output:} Latent map $z^s_0$, edited latent map $z_0$ corresponding to $P_s$ and $P$.
        \begin{algorithmic}[1]
            \FOR{$t = T, T-1, ..., 1$}
                \STATE $\epsilon_s, \{Q_s, K_s, V_s\}\leftarrow \epsilon_\theta(z^s_t, P_s, t)$;
                \STATE $z^s_{t-1} \leftarrow \text{Sample}(z^{s}_{t}, \epsilon_s)$;
                \STATE $\{Q, K, V\} \leftarrow \epsilon_\theta(z_t,P,t)$;
                \STATE $\{Q^*, K^*, V^*\} \hspace{-1mm}\leftarrow \text{EDIT}(\{Q,K,V\}, \{Q_s, K_s, V_s\})$;
                \STATE $\epsilon = \epsilon_\theta(z_t, P, t; \{Q^*, K^*, V^*\})$;
                \STATE $z_{t-1} \leftarrow \text{Sample}(z_{t}, \epsilon)$;
            \ENDFOR
        \end{algorithmic}
        \label{alg:masactrl}
        \textbf{Return} $z^s_0, z_0$
\end{algorithm}

\begin{figure}[th]
    \centering
    \includegraphics[width=0.99\linewidth]{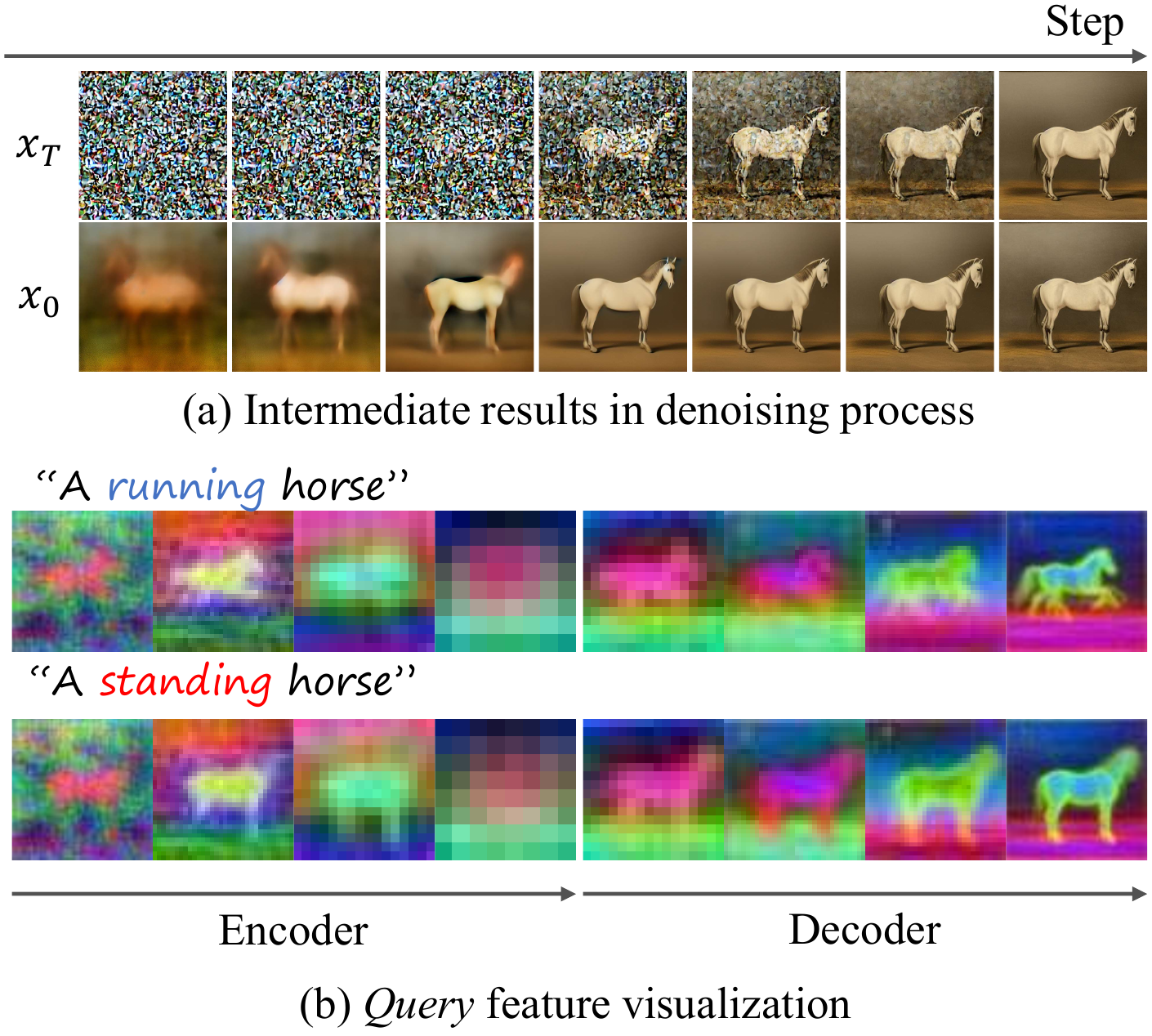}
    \caption{(a) The intermediate results during the iterative denoising process, and (b) visualization of the projected \textit{Query} features $Q$ in the self-attention layers of the U-Net at the 15th sampling step.}
    \label{fig:intermediate_vis}
\end{figure}

However, intuitively performing such attention control on all layers among all denoising steps will result in an image $I$ that is nearly the same as the reconstructed image $I_s$. We argue the reason is that performing self-attention control in the early steps can disrupt the layout formation of the target image. In the premature step, the target image layout has not yet been formed~(shown in Fig.~\ref{fig:intermediate_vis}(a)), and we further observe the Query features in the shallow layers of U-Net~(\eg, encoder part) cannot obtain clear layout and structure corresponding to the modified prompt~(shown in Fig.~\ref{fig:intermediate_vis}(b)). Thus we cannot obtain the image with the desired spatial layout.

Therefore, we propose to control the mutual self-attention \textit{only in the decoder part of the U-Net after several denoising steps}, due to the formed clear target image layout and semantically similar features (see Fig.~\ref{fig:intermediate_vis}). We can change the original layout into the target one with edited prompt $P$ and keep the main objects unchanged with proper starting denoising step $S$ and layer $L$ for synthesis and editing. Thus the EDIT function in Alg.~\ref{alg:masactrl} can be formulated as follows:
\begin{equation}
    \label{eq:masactrl_edit}
    \text{EDIT} := \left \{
    \begin{aligned}
        & \{Q, K_s, V_s\},\quad \text{if}\quad t > S \quad \text{and} \quad l > L, \\
        & \{Q, K, V\},\quad \text{otherwise},
    \end{aligned}
    \right.
\end{equation}
where $S$ and $L$ are the time step and layer index to start attention control, respectively. 

In the early steps, the composition and the shape of the object can be roughly generated complying with the target prompt $P$. Then the content information from the source image $I_s$ is queried by the mutual self-attention mechanism to fill the generated layout of $I$. After iterative denoising, we can obtain the synthesized image with similar contents in the source image and structure of $I^*$ that conforms to the input prompt. Note that our algorithm does not require finetuning or optimization, bringing many conveniences for ordinary users for content creation.

\subsection{Mask-Guided Mutual Self-Attention} \label{sec:mask-guided}
We also observed the above synthesis/editing would fail since the object and background are too similar to be confused. 
To tackle this problem, one feasible way to is to segment the image into the foreground and the background parts and query contents only from the corresponding part. 
Inspired by previous work~\cite{hertz2022prompt, tang2022daam}, the cross-attention maps correlating to the prompt tokens contain most information of the shape and structure. 
Therefore, we utilize the semantic cross-attention maps to create a mask to distinguish the foreground and background in both source and target images $I_s$ and $I$. 

Specifically, at step $t$, we first perform a forward pass with the fixed U-Net backbone with prompt $P_s$ and edited prompt $P$, respectively, to generate intermediate cross-attention maps.
Then we average the cross-attention maps across all heads and layers with the spatial resolution $16 \times 16$. The resulting cross-attention maps are denoted as ${A^c_t} \in \mathbb{R}^{16\times 16 \times N}$, where $N$ is the number of the textual tokens. 
We then obtain the averaged cross-attention map for the token correlated to the foreground object. We denote $M_s$ and $M$ as masks extracted for the foreground objects in $I_s$ and $I$, respectively. With these masks, we can restrict the object in $I$ to query contents information only from the object region in $I_s$:

\begin{figure*}[!t]
    \centering
    \includegraphics[width=0.99\linewidth]{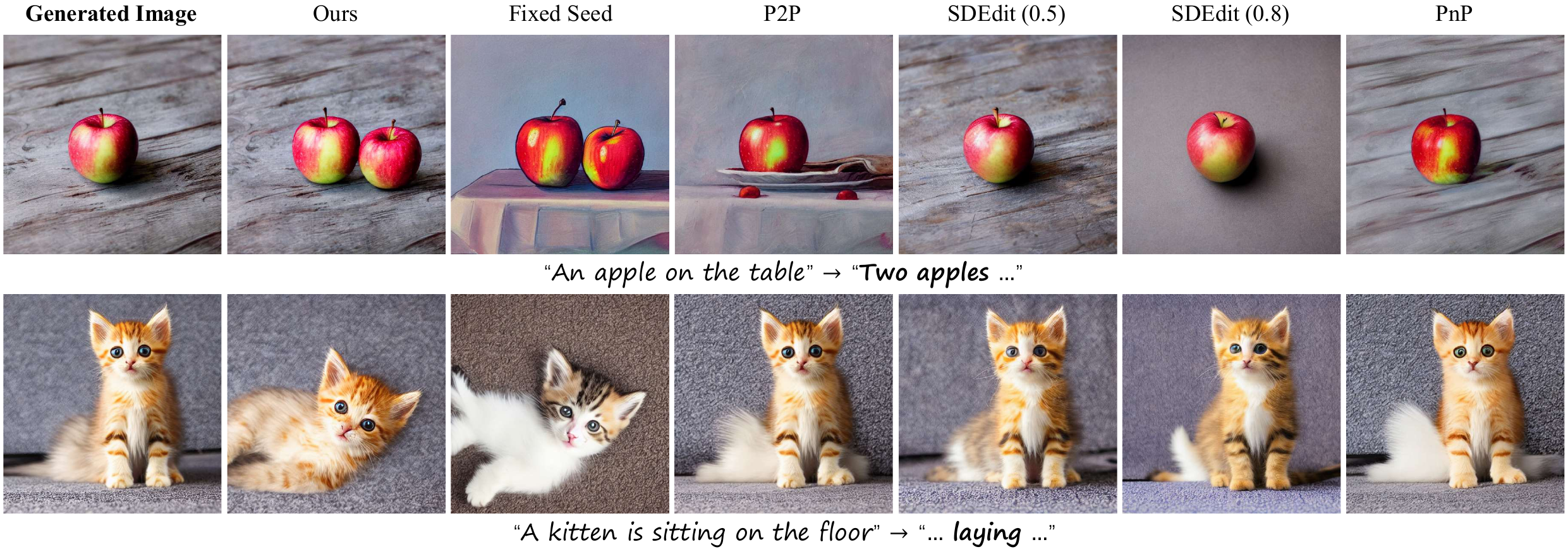}
    \caption{Synthesis results of different methods on the synthetic images. Our method enables consistent synthesis by combining the layout of the target prompt and the contents of source generated image. From left to right: the source generated source image with source prompt, synthesis results with the proposed MasaCtrl method, synthesis results from target prompt with the same random seed of source image, synthesis results with existing methods    P2P~\cite{hertz2022prompt}, SDEdit~\cite{meng2021sdedit}, and PnP~\cite{tumanyan2022plug}. }
    \label{fig:results_synthetic}
\end{figure*}

\begin{figure*}[!t]
    \centering
    \includegraphics[width=0.99\linewidth]{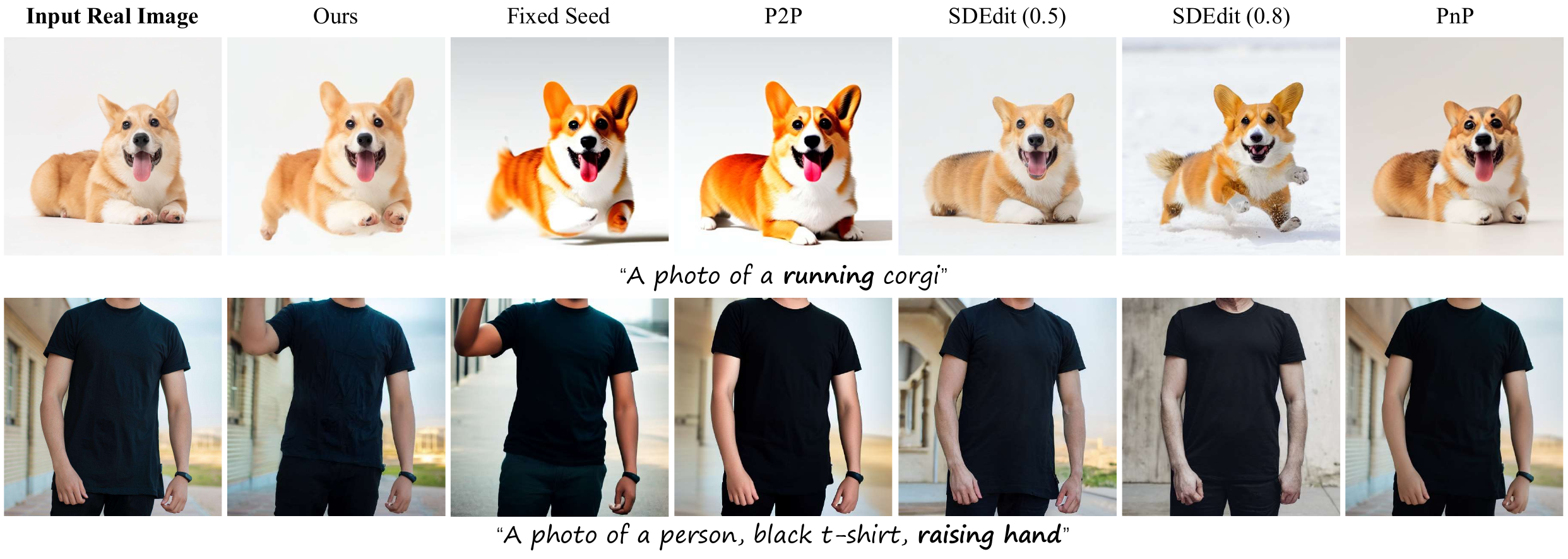}
    \caption{Real image editing results of different editing methods on real images. From left to right: the input real image, synthesis results with the proposed MasaCtrl method, synthesis results from target prompt with the same random seed of source image, synthesis results with existing methods    P2P~\cite{hertz2022prompt}, SDEdit~\cite{meng2021sdedit}, and PnP~\cite{tumanyan2022plug}. }
    \label{fig:results_real}
\end{figure*}

\begin{align}
    \label{eq:mask_attn}
    f^l_{o} &= \text{Attention}(Q^l, K^l_{s}, V^l_{s}; M_s), \\
    f^l_{b} &= \text{Attention}(Q^l, K^l_{s}, V^l_{s}; 1 - M_s), \\
    \bar{f}^l &= f^l_{o} * M + f^l_{b} * (1 - M),
\end{align}
where $\bar{f}^l$ is the final attention output.
The object region and the background region query the content information from corresponding restricted areas rather than all features.

\begin{figure*}
    \centering
    \includegraphics[width=\linewidth]{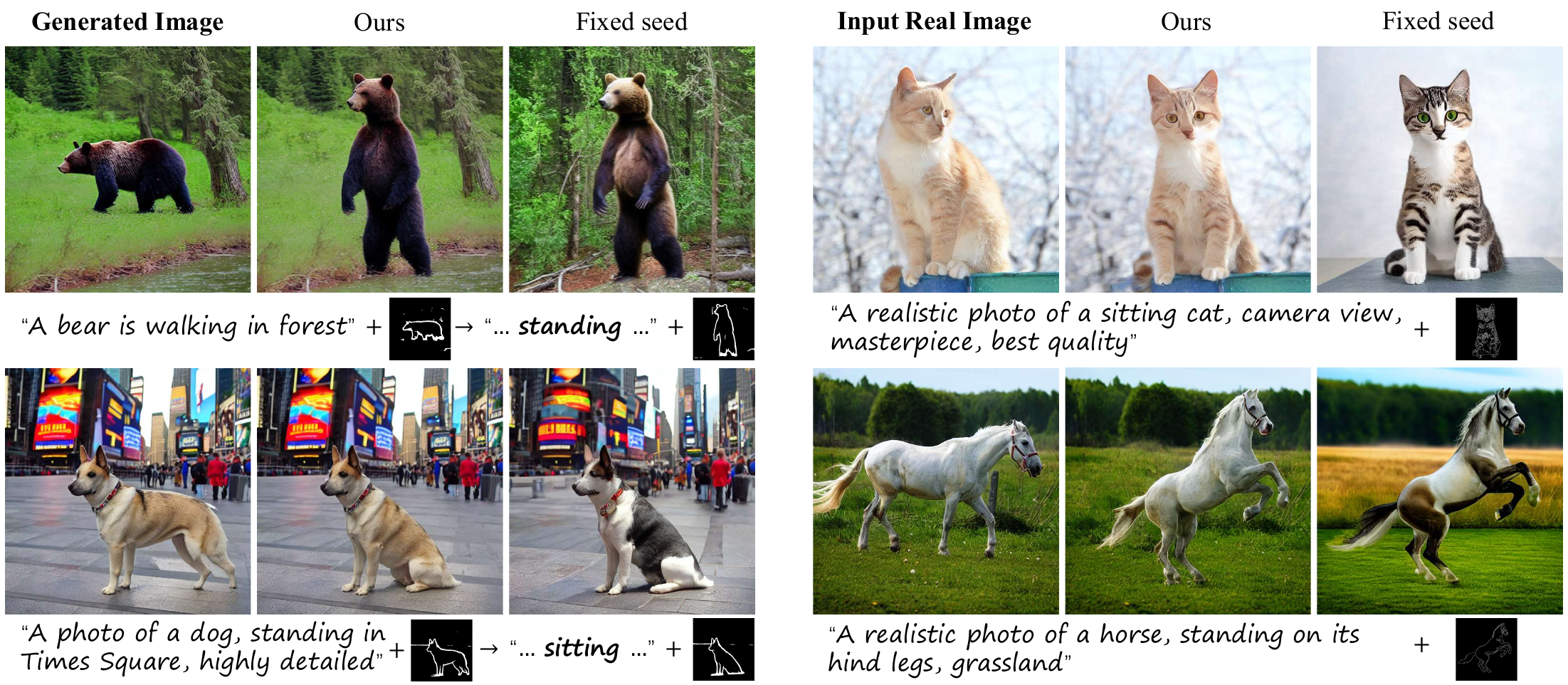}
    \caption{Consistent synthesis results~(left part) and real image editing results~(right part) with MasaCtrl integrated into T2I-Adapter~\cite{mou2023t2i}. }
    \label{fig:results_adapter}
\end{figure*}

\begin{figure*}
    \centering
    \includegraphics[width=\linewidth]{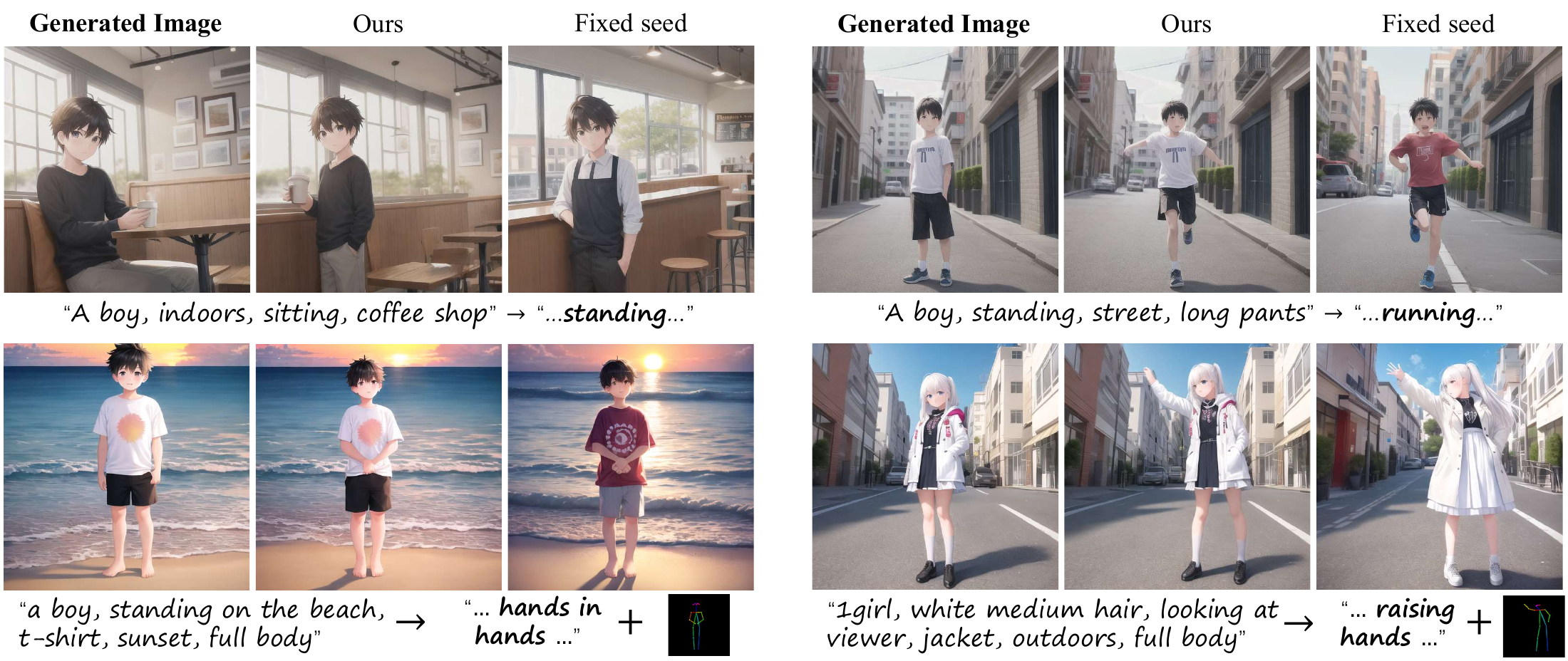}
    \caption{Synthesis results with Anything-V4 checkpoint. We see that consistent images can be synthesized by directly modifying the text prompts with the proposed MasaCtrl. }
    \label{fig:results_angthing}
\end{figure*}

\begin{figure*}
    \centering
    \includegraphics[width=\linewidth]{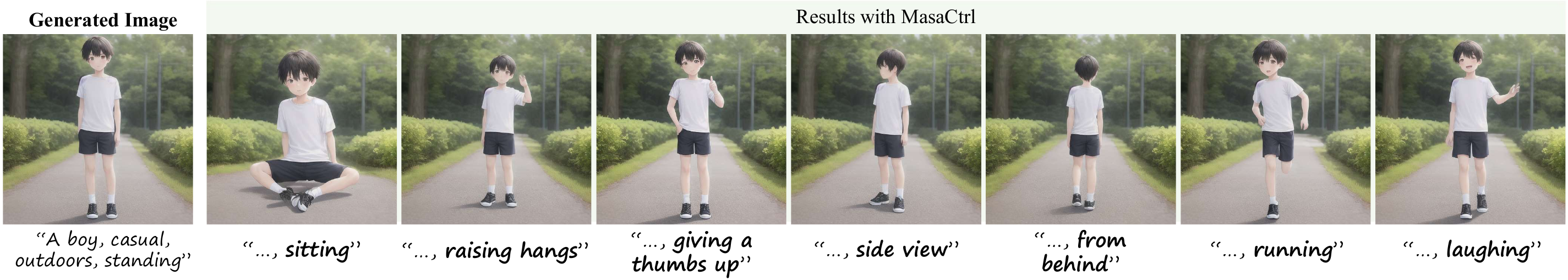}
    \caption{Multiple consistent synthesis results with proposed MasaCtrl on Anything-V4 checkpoint. }
    \label{fig:coherent_synthesis}
\end{figure*}

\begin{figure*}
    \centering
    \includegraphics[width=\linewidth]{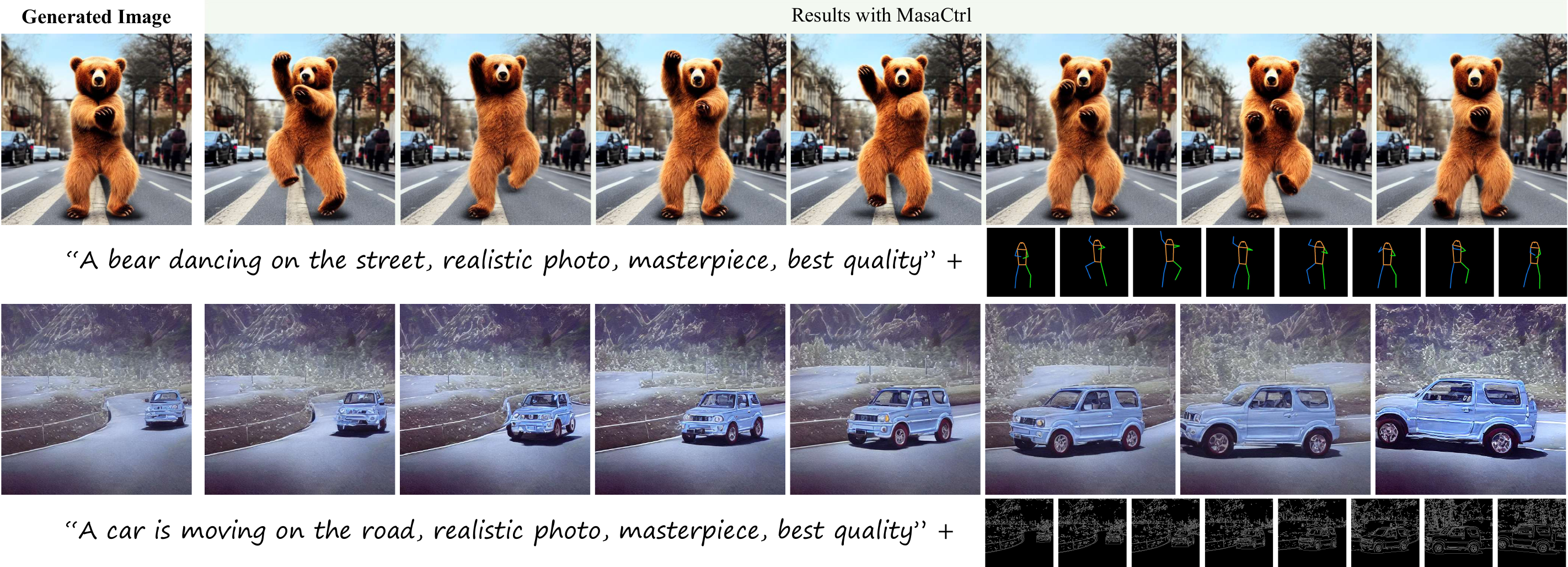}
    \caption{Video synthesis results of proposed MasaCtrl with T2I-Adapter~\cite{mou2023t2i}~(key-pose and canny guidance). }
    \label{fig:coherent_synthesis_adapter}
\end{figure*}

\subsection{Integration to Controllable Diffusion Models} \label{sec:controllable}

Our method can be easily integrated into existing controllable image synthesis method~(\eg, T2I-Adapter~\cite{mou2023t2i} and ControlNet~\cite{zhang2023adding}) for more faithful non-rigid image synthesis and editing.
These methods enable more controllable~(\eg, pose, sketch, segmentation map) image synthesis to the original Stable Diffusion model. Thus we can use them to synthesize images with desired poses, shapes, and views. Nevertheless, they still cannot synthesize images with similar contents in a reference source image. Since our method can query image contents~(\eg, textures, color) from a reference image, we can easily integrate our approach into these models to generate more coherent images without fine-tuning the SD model or optimizing the textual embedding~\cite{kawar2022imagic}. 

Specifically, we follow the same process shown in Alg.~\ref{alg:masactrl}, and the desired target image synthesis process changes from the original SD model to using these controllable models instead. In the following experiment part, we demonstrate the effectiveness of such a combination that can synthesize images coherently.

\section{Experiments}
\noindent\textbf{Setup.}
We apply the proposed method to the state-of-the-art text-to-image Stable Diffusion~\cite{rombach2022high} model with publicly available checkpoints v1.4. We also validate the proposed method on the pre-trained anime-style model Anything-V4. Meanwhile, we perform editing on both synthetic images and real images. For real image editing, we first invert the image into  the initial noise map with DDIM deterministic inversion~\cite{song2020denoising}. Note that we set the starting noise map the same for source prompt $P_s$ and the desired prompt $P$ unless otherwise specified. During sampling, we perform DDIM sampling~\cite{song2020denoising} with 50 denoising steps, and the classifier-free guidance is set to 7.5. The step and layer to start attention control is set to $S=4, L=10$ as default. Note that it may be changed for specific checkpoints.

\subsection{Comparisons with Previous Works}
We mainly compare the proposed tuning-free method to the current prompt-based editing methods with diffusion models, including tuning-free methods SDEdit~\cite{meng2021sdedit}, P2P~\cite{hertz2022prompt}, PnP~\cite{tumanyan2022plug}. We use their open-sourced codes to produce the editing results~\footnote{SD model utilizes SDEdit for img2img synthesis. Thus, we directly utilize the script in SD to synthesize results.}. 

The synthesis results are shown in Fig.~\ref{fig:results_synthetic}. By directly modifying the text prompt, our method can synthesize content-consistent images. These synthesized images (1) contain contents (foreground objects and background) that are highly similar to those in the generated source images (first column  of Fig.~\ref{fig:results_synthetic}) and (2) highly comply with the target prompt $P$ (shown in the third column of Fig.~\ref{fig:results_synthetic}). While existing methods fail to synthesize desired images conforming to the target text prompt.

Our method also achieves good results in \textbf{editing real images} (shown in Fig.~\ref{fig:results_real}). Note that performing non-rigid editing on real images is very challenging. 
We see that the contents~(foreground objects and background) in the images edited by the proposed method are much similar to the input image. These results demonstrate the effectiveness of the proposed method. Meanwhile, unlike the previous method Imagic~\cite{kawar2022imagic} that requires finetuning the network, optimizing the textual embedding, and interpolation between the optimized embedding for a single edit, our tuning-free method provides the users a simple and convenient way for non-rigid image editing.

We further analyze the reasons attributed to the failure of existing methods (\ie, P2P, SDEdit, and PnP). These methods try to keep the original layout or object shape and pose unchanged by leveraging the layout information encoded in the cross-attention maps~(P2P), features~(PnP), and original input images~(SDEdit). Meanwhile, the contents in the formed image mainly come from the encoded text embedding. As a result, the images synthesized by these methods have similar layouts to the source image but have different contents. 
In our proposed method MasaCtrl, the structure of the desired image is first determined by the former iteration in the denoising process, and the final image is formed by obtaining the image content from the source image. Thus we can perform various types of non-rigid image editing.

\begin{figure*}[!t]
    \centering
    \includegraphics[width=0.99\linewidth]{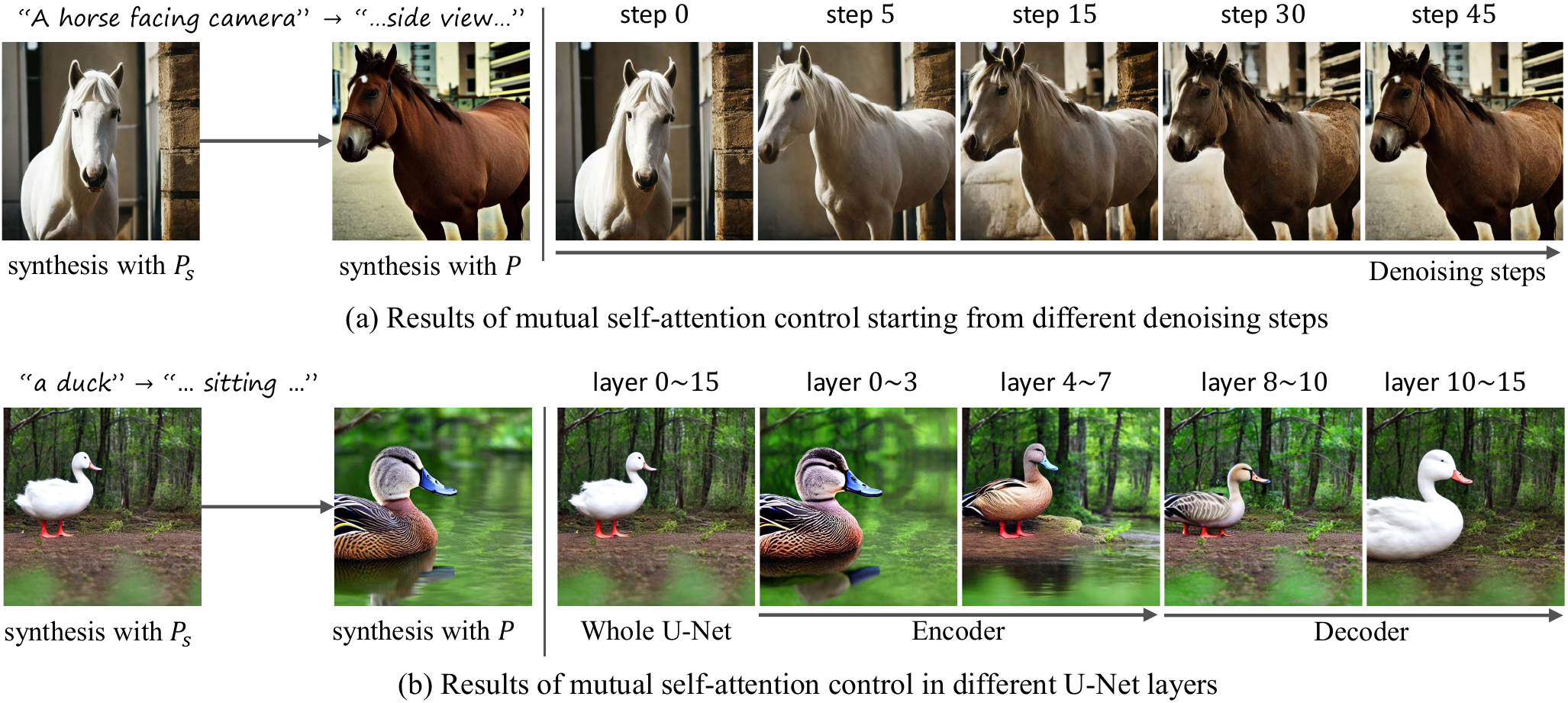}
    \caption{Results of mutual self-attention control in different denoising steps~(a) and different U-Net layers~(b). We see that only performing mutual self-attention control after several denoising steps~(\eg, step 5), and in the decoder part (\eg, layer $10\sim 15$) can preserve the shape and structure information from target prompt $P$ and query contents from the source image with prompt $P_s$.}
    \label{fig:step_layer_analysis}
\end{figure*}

\subsection{Results with T2I-Adapter}
The initial layout controlled by modifying the text prompt usually fails due to the inherent drawbacks of the Stable Diffusion model. Therefore, we further integrate our method into existing controllable synthesis pipelines to obtain stable synthesis and editing results. The synthesis and real image editing results with the T2I-Adapter~\cite{mou2023t2i} are shown in Fig.~\ref{fig:results_adapter}. We see that T2I-Adapter can generate an image with desired target layout, yet with different contents in the source image. While our method can effectively combine the layout synthesized by T2I-Adapter with the target prompt and contents in the source image. Therefore, more faithful and fine-grained synthesis and editing results can be obtained. Note that we may change the attention strategy of starting denoising step $S$ and U-Net layer $L$ to obtain faithful results ($S=2$, $L=8$ in our experiment), since the target layout is strongly controlled by extra guidance (thus we can perform attention control in early steps and layers to query faithful contents in the source image, refer to the ablation study Sec.~\ref{sec:ablation_study} for the analysis). 

\subsection{Robustness to Other Models: Anything-V4}

We also apply our method to the domain-specific models, \ie, the amine-style model Anything-V4. Fig.~\ref{fig:results_angthing} shows the synthesis results of our method and the model with fixed random seeds. The proposed method MasaCtrl can faithfully synthesize images while preserving the object identity and background in original anime-style images, further demonstrating the generalizability of the proposed method. 
Meanwhile, we further perform consistent image synthesis in Fig.~\ref{fig:coherent_synthesis}. We can control the pose and action, even expression, with the proposed method by directly modifying the text prompt, demonstrating the consistent synthesis capability of MasaCtrl.

\subsection{Extension to Video Synthesis}
Although our method is designed for image synthesis and editing,
we can easily extend our method for the video synthesis task using T2I-Adapter and ControlNet with a series of dense coherent guidance (\eg, pose, edge). Specifically, we first generate a source image with the prompt $P$ and guidance~(shown in the first column of Fig.~\ref{fig:coherent_synthesis_adapter}). In the following frame synthesis process, we perform MasaCtrl on each frame to generate the target frame that is content-consistent with the source frame. Fig.~\ref{fig:coherent_synthesis_adapter} shows the video synthesis results with coherent dense guidance. The proposed method successfully synthesizes consistent frames with highly similar content (please visit the project page for more video results). However, our method can only animate the foreground objects (such as the bear in Fig.~\ref{fig:coherent_synthesis_adapter}) and hardly bring the background alive. Therefore, video-based approaches still need to be explored since the current MasaCtrl has a significant limitation in synthesizing scenes with background dynamics.

\subsection{Ablation Study}\label{sec:ablation_study}
The results of both synthetic image synthesis and real image editing can demonstrate the effectiveness of the proposed mutual self-attention. We further analyze the control strategy in terms of different starting steps at the denoising process and the layers in the denoising U-Net. From Fig.~\ref{fig:step_layer_analysis}(a), we see that performing mutual self-attention in the premature step can only synthesize an image identical to the source image, conveying all source image contents and ignoring the layout from the target prompt. As the step increases, the synthesized desired image can maintain the layout from the target prompt and the contents from the source image. While the image would gradually lose the source image contents and eventually becomes the image synthesized images without mutual self-attention control. We also observe a similar phenomenon when performing control in different U-Net layers shown in Fig.~\ref{fig:step_layer_analysis}(b). Performing control among all layers can only generate an image identical to the source image. Performing control in low-resolution layers (\eg, layer $4\sim 10$) cannot preserve the source image contents and target layouts. While in high-resolution layers (\eg, layer $0\sim 3$, $10\sim 15$), the target layout can be maintained, and the source image contents can only be transformed when controlled in the decoder part. As a result, the proposed method performs control in the decoder part of U-Net after several denoising steps.

\begin{figure}[b]
    \centering
    \includegraphics[width=\linewidth]{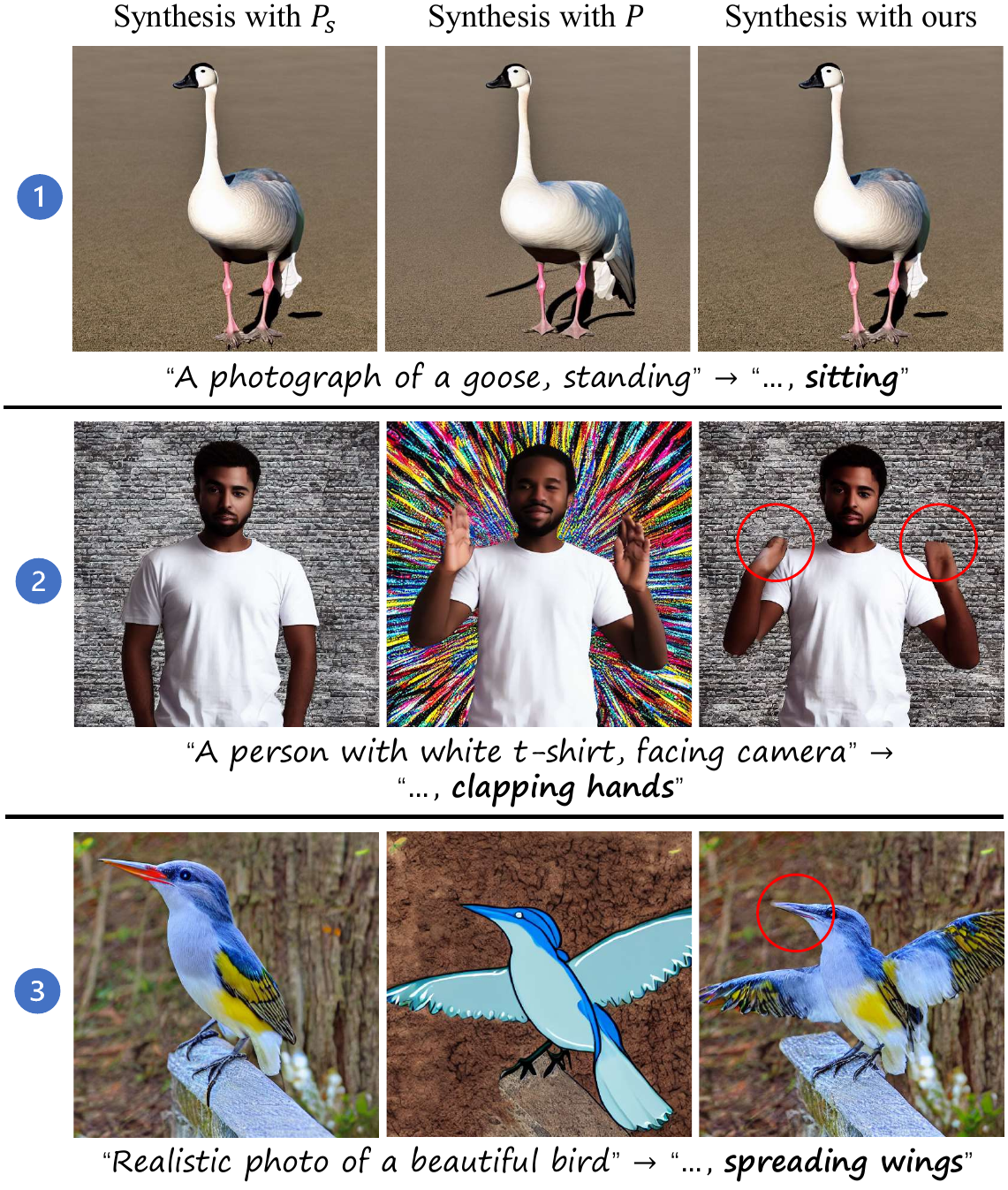}
    \caption{Different types of failure cases. }
    \label{fig:failure_case}
\end{figure}

\section{Limitations and Discussion}
Our method inherits most of the limitations of Stable Diffusion in generating desired images, and suffers from the following main aspects. First, since our method heavily relies on the image layout synthesized from the target prompt $P$, it would fail if the SD model could not generate a desired layout or shape, as shown in Fig.~\ref{fig:failure_case}(1). Although recently proposed controllable strategies~\cite{mou2023t2i, zhang2023adding} can alleviate this on the pre-trained SD model with various guidance, it still may fail. In addition, even if the SD model can generate the corresponding image layout, our method will fail when the target image contains unseen content or the target image layout/structure changes drastically. As shown in Fig.~\ref{fig:failure_case}(2), the SD model can synthesize the target layout that complies with the target prompt $P$ while with different contents~(\ie, the identity of the person and the background) from the source image. MasaCtrl can generate an image consistent with the source image but suffer from the artifact~(the palm marked by the red circle). This is because the source image does not contain any contents related to the palm, thus the desired image cannot query the contents of the palm. Meanwhile, as shown in Fig.~\ref{fig:failure_case}(3), although our method can synthesize desired image that is highly similar to the source image, we also found there still are some slight differences~(the color of the bird's beak marked by the red circle) between the source image and the edited image. How to tackle these problems is our future work. 

\section{Conclusion}
We propose MasaCtrl, a tuning-free mutual self-attention control method applied to T2I diffusion models for non-rigid consistent image synthesis and editing. We convert the self-attention mechanism in diffusion models into a cross one, dubbed mutual self-attention, enabling effective structure and appearance query from the source image when applied to specific denoising steps and U-Net layers. We further consider the confusion problem of the foreground objects and background during the querying process and alleviate it with a mask-guided strategy. Meanwhile, our method can be easily integrated into recently proposed controllable strategies over diffusion models and perform consistent image synthesis and editing without finetuning the model and textural embedding. We believe such a method provides ordinary users with a convenient and effective way for content creation under text description.

\clearpage

{\small
\bibliographystyle{ieee_fullname}
\bibliography{egbib}
}

\end{document}